# Lost in Translation: Policymakers are not really listening to Citizen Concerns about AI

## Susan Ariel Aaronson and Michael Moreno[1]

**Executive Summary**

The world's people have strong opinions about artificial intelligence (AI), and they want policymakers to listen. Policymakers are seeking public comment on AI. But as they translate public comments into policy, much of what citizens have to say is getting lost. As a result, policymakers are missing an important opportunity to build public trust in AI and its governance.

This paper compares three countries that sought public comment on AI risks and policies to mitigate those risks. The authors used a landscape analysis to examine how Australia, Colombia, and the US formally invited citizens to comment on AI risk and governance. However, we did not find that citizens and policymakers had established a constructive dialogue about AI. First, the three countries did not go out of their way to obtain diverse public comments. They could have done more to ensure that their citizens knew about the call and gave these constituents the information they needed to respond to the calls (a supply-side marketing problem). In each nation, less than one percent of the population responded to the call. Thirdly, policymakers did not appear responsive to what they heard from the few who participated. These officials were unwilling and unable to create a feedback loop. In short, we found a persistent gap between the promise and practice of participatory AI governance.

The authors believe that the current process in all 3 countries is unlikely to build trust and legitimacy in AI or AI governance because policymakers are not listening and responding to public concerns. The authors made 8 recommendations to modernize this process including: supporting AI literacy; consistently monitoring public comment; widely market calls for comment; regularly hold online town halls on AI; utilize innovative engagement strategies; ensure participation from underrepresented groups, be responsive to what people say and do more to ensure that their citizens have an opportunity to comment.

[1] This paper is based on work supported in part by the NSF-NIST Institute for Trustworthy AI, which is supported by the National Science Foundation under award 2229885. Any opinion, finding, and conclusion or recommendation expressed herein represents that of the authors and does not necessarily reflect the views of NSF.

**Introduction**

Democratic governance is a dialogue between the government and its constituents (Dewey: 1927). As democracies evolved in the 19th and 20th centuries, democratic policymakers understood that they needed a mechanism to inform their citizens about changing policies and government opportunities. They could then use these comments to improve governance. In so doing, democracies could build trust and legitimacy. They may also achieve better public policies (OECD: 2022).

Like many of their constituents, policymakers struggle to understand and govern AI. Because AI is rapidly evolving, government officials tend to lean on those individuals (academics and business leaders) who have developed or invested in AI. In turn, these experts often have a direct interest in whether and how AI should be regulated (Aaronson: 2025; OECD: 2025a). Often, these experts work for or are affiliated with the major suppliers of AI. Meanwhile, these AI suppliers are the same data giants that have captured much of the world's personal data online through their provision of online services such as social networks (Zuboff: 2019, and Bradford: 2024). These firms already have significant political influence due to their economic power. For example, in many countries, they have pushed back on digital taxes, AI regulation, and laws requiring them to support traditional media (Carvao et al. 2025; Clement: 2023).

Scholars of technology believe that consultation is particularly important because of the challenging nature of AI. AI issues are complex and require citizens to be well-informed (AI literacy). In turn, policymakers must be willing and able to devote time and effort to ensuring their constituents have the information they need to comment (Maas: 2023; Aaronson and Zable: 2023).

Moreover AI, like many technologies, is simultaneously beneficial and risky. As example, AI use may improve economic growth and improve human welfare but also create risks to individuals and society as a whole (OECD: 2024, 2025). Given the Janus nature of AI, the OECD recommends that "engaging diverse actors early in the technology development cycle enriches the understanding of issues, fosters trust, and aligns technological innovation with societal needs." Officials should take care "to balance the range of perspectives and ensure that vocal vested interests do not dominate the process. Tools for societal engagement, including capacity-building, communication, consultation and co-creation should be leveraged to ensure broad-based participation and alignment of science and co-design of technology strategies and governance" (OECD: 2025a, p. 11).

Meanwhile, polls reveal that citizens are deeply concerned about AI. A 2023 survey of 24,000 people in 21 countries by the Schwartz Reisman Institute of the University of Toronto found that some 73% of those polled understood what AI is, but they were less well-informed about specific applications. In general, the public is concerned about potential risks to employment, human autonomy, democracy, and human rights (Schwartz-Reisman Institute for Technology and Society: 2024). Another 2023 global study by IPSOS found that from 2022 to 2023, the proportion of those who think AI will dramatically affect their lives in the next three to five years increased from 60% to 66% (Standford HCAI, 2024). When asked who should regulate AI, the pollsters found that people prefer that the AI developing companies should do so. But when asked if they trust tech companies to self-regulate, only 21% of the respondents said yes (Schwartz-Reisman Institute for Technology and Society: 2024).

This is the third paper in a series of four papers for CIGI, examining the feedback loop between policymakers and their constituents about AI. In this paper, the authors compare how three governments—the United States, Colombia, and Australia—sought to engage citizens in AI governance through formal public consultation processes. We chose these countries for 3 reasons: first, acknowledged and openly published all the comments they received; some governments such as Canada do not for privacy reasons. Secondly, the countries of choice public come from different regions of the world. Thirdly, we had not done a comparative analysis of them in an earlier paper3333 Herein, the authors analyze the U.S. National Telecommunications and Information Administration's (NTIA) AI Open Model Weights Request for Comment,[2] Colombia's Ministry of Science, Technology and Innovation (MinCiencias) consultation on the roadmap for ethical and sustainable AI adoption, and Australia's Department of Industry, Science, and Resources' (DISR) "Supporting Responsible AI" discussion paper. As before, we used a landscape analysis to plot government actions and citizen responses. As with our earlier analyses we have five key questions: How and when the governments reached out to the public for comment; Who participated in the call for comment; What materials did the government provide to prepare the public to give informed advice; Did policymakers make efforts to ensure a broad cross-section of people knew about and could comment on the proposed policy; Did the governments provide evidence that it made use of the feedback it received?

Our findings reveal that all 3 governments did not receive comments from a sizable and diversified cross-section of their constituents. In fact, Australia received 510 comments: the US 326 comments, and Colombia 73 (way less than 01% of their populace). While the governments differed in how they sought comments, none of them widely marketed the call to their constituents using a wide range of platforms and tactics to attract a broad response. All three provided some form of background material to guide their constituents. However, policymakers did not make an effective effort to ensure that their citizens had the information they needed to comment. In sum, public comment got lost in translation or was never translated into policy. The authors conclude that if policymakers genuinely want to achieve a participatory approach to AI governance, they need to rethink the consultative process for the age of AI.

[2] This was the topic of another CIGI paper—Talking to a Brick Wall, CIGI Paper No. 219, 20, where we examined the consultation in depth. Here we compare the 3 countries approaches.

The paper proceeds as follows: we begin with an overview of our methodology, next discuss the traditional process of seeking input. The authors then examine our 5 questions in depth. The authors then conclude with some suggestions for policymakers.

**Methodology**

In order to compare and characterize countries the authors have established a consistent methodology for their analysis over the past few years. Specifically, the authors relied on tools delineated by the International Association for Political Participation (IAP2) to describe the interaction between the three countries and the public on this issue. The IAP2 is an international association that provides public participation practitioners around the world with the tools, skills, and networking and training opportunities to advance and extend the practice of public participation. It has published both a set of core values and a spectrum delineating the levels of public participation in a democracy.

## IAP2 Spectrum of Public Participation

iap2 international association for public participation

IAP2's Spectrum of Public Participation was designed to assist with the selection of the level of participation that defines the public's role in any public participation process. The Spectrum is used internationally, and it is found in public participation plans around the world.

INCREASING IMPACT ON THE DECISION

| | INFORM | CONSULT | INVOLVE | COLLABORATE | EMPOWER |
|---|---|---|---|---|---|
| PUBLIC PARTICIPATION GOAL | To provide the public with balanced and objective information to assist them in understanding the problem, alternatives, opportunities and/or solutions. | To obtain public feedback on analysis, alternatives and/or decisions. | To work directly with the public throughout the process to ensure that public concerns and aspirations are consistently understood and considered. | To partner with the public in each aspect of the decision including the development of alternatives and the identification of the preferred solution. | To place final decision making in the hands of the public. |
| PROMISE TO THE PUBLIC | We will keep you informed. | We will keep you informed, listen to and acknowledge concerns and aspirations, and provide feedback on how public input influenced the decision. | We will work with you to ensure that your concerns and aspirations are directly reflected in the alternatives developed and provide feedback on how public input influenced the decision. | We will look to you for advice and innovation in formulating solutions and incorporate your advice and recommendations into the decisions to the maximum extent possible. | We will implement what you decide. |

Table #1 reprinted by permission from IAP2

Our questions remain consistent:

1. How and when did the government engage with its citizens?
2. What materials did the government provide to prepare the public to give informed advice?
3. Did policymakers attempt to ensure a broad cross-section of people knew about and could comment on the proposed policy?
4. Who participated?
5. Did the government provide evidence that it made use of the feedback it received?

Then the authors used the IAP2 spectrum to characterize each case study (Aaronson and Zable: 2023, p. 14)

To assess each country's call for comments and response, the researchers began by creating a list of everyone who responded to the call. They then conducted a landscape analysis, dividing the respondents into groupings that reflected their own descriptions of themselves or their organizations. Readers can view the analysis as follows: Overview of the February 2024 Request for Comment on dual-use foundation models l;[3] Colombia's "Hoja de Ruta para garantizar la adopción ética y sostenible de la IA,"[4] and the 2023 "Supporting Responsible AI" discussion paper from the government's consultation hub and likewise compiled them into an Excel sheet.[5] The findings and background data will be available at the Digital Trade and Data Governance Hub research website.[6]

**A Historical Perspective on the Process of Seeking Input**

Our three case studies developed similar processes to inform and speak with their constituents. For example, in 1935, the US Congress passed a law creating the Federal Register. It listed Presidential proclamations and Executive orders; documents required to be published by Act of Congress; and documents authorized to be published by regulations (National Archives: 1936). Since 2003, US agencies have used Regulations.gov as their principal vehicle for soliciting public comments on proposed regulations and storing relevant background information (Cogliese: 2001). Australia's Parliament created the Federal Register of Legislation (the Register) in 2023. It contains the full text and details of the lifecycle of individual laws and the relationships between them. Colombia conducts its consultations via Google Forms and uses social media and its official websites to market these requests (MinCiencias 2024b; MinCiencias 2024e; MinCiencias 2024b).

But policymakers in the US and other democracies gradually recognized that they had established only one-way communication. Their strategies allowed them to inform, but not to receive feedback. In 1946, the US Congress passed the Administrative Procedures Act, which required that Agencies not only publish notices of proposed rulemaking in the Federal Register, but also provide an opportunity for public comment before final rules could be put into effect. Australia relies on a slightly different process. The government often creates discussion papers and asks for citizen to respond to these papers. (Australian Government: 2020, 12-13).[7] Colombia formally established mechanisms for public consultation through Article 103 of its 1991 Constitution (Asamblea Nacional Constituyente, 1991), which recognizes various forms of citizen participation, and further institutionalized these tools with Law 134 of 1994, which created the National Popular Consultation as a means for the President to seek public input on issues of national significance (Congreso de Colombia, 1994).

[3] NTIA Spreadsheet and RFI: Spreadsheet; RFI
[4] Colombia Spreadsheet and Consultation: Spreadsheet; Consultation
[5] Australia Spreadsheet and Consultation: Spreadsheet; Industry Consultation
[6] https://datagovhub.elliott.gwu.edu/

By the 21st century, it was clear that this approach of ‘asking for public comment ‘was no longer effective. First, the few people who respond tend to be policy insiders, those who closely follow governance. These respondents already have a voice in their country’s capital and are sophisticated in influencing policy. Secondly, simply calling for public comment cannot guarantee a diverse and broad sample of the populace. Third, technologies such as the Internet and data driven technologies such as artificial intelligence were changing rapidly, and policymakers could not keep pace. So, the OECD suggested that governments adopt a new approach to regulation. Agile governance or regulation “involves more holistic, open, inclusive, adaptive and better coordinated governance models to enhance systemic resilience by enabling the development of agile, technology neutral and adaptive regulation that upholds fundamental rights, democratic values and the rule of law.”[8] But the OECD did not explain how governments could achieve that balance.

**Findings**

This section provides summaries of our findings:

**1. Who responded to the call for comment?**

[7] https://www.legislation.gov.au/; and https://www.homeaffairs.gov.au/reports-and-publications/submissions-and-discussion-papers#:~:text=%E2%80%8B%E2%80%8B%E2%80%8B%E2%80%8B%E2%80%8B%E2%80%8B%E2%80%8B%E2%80%8B%E2%80%8B%E2%80%8B%E2%80%8BCyber%20security%20legislative%20reforms%20The%20d%E2%80%8Bepartment%20seeks,Minister%20for%20Emergency%20Management%2C%20Senator%20the%20Hon

[8] Recommendation of the Council for Agile Regulatory Governance to Harness Innovation, OECD/LEGAL/0464, Adopted on: 06/10/2021 https://legalinstruments.oecd.org/en/instruments/OECD-LEGAL-0464

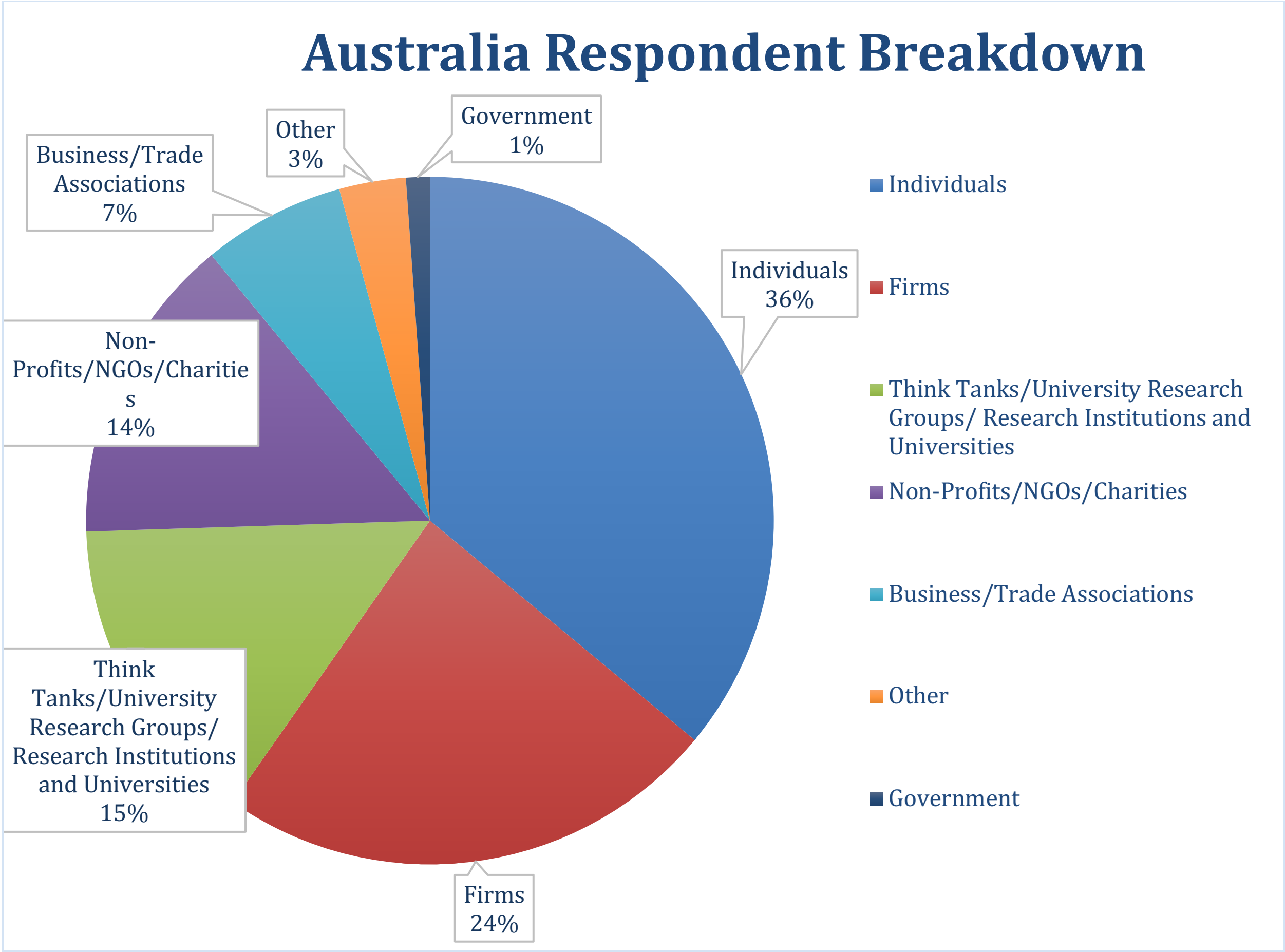

*Chart #1* by Michael Moreno

| Australia Respondent Breakdown | | |
|---|---|---|
| **Respondent** | **% of Total** | **Count** |
| Individuals | 36% | 161 (47 anonymous) |
| Firms | 23.70% | 106 |
| Think Tanks/University Research Groups/ Research Institutions and Universities | 14.80% | 66 |
| Non-Profits/NGOs/Charities | 14.30% | 65 |
| Business/Trade Associations | 6.70% | 30 |
| Other | 3.10% | 14 |
| Government | 1.30% | 5 |
| **Total** | | **447** |

Table #2 by Michael Moreno

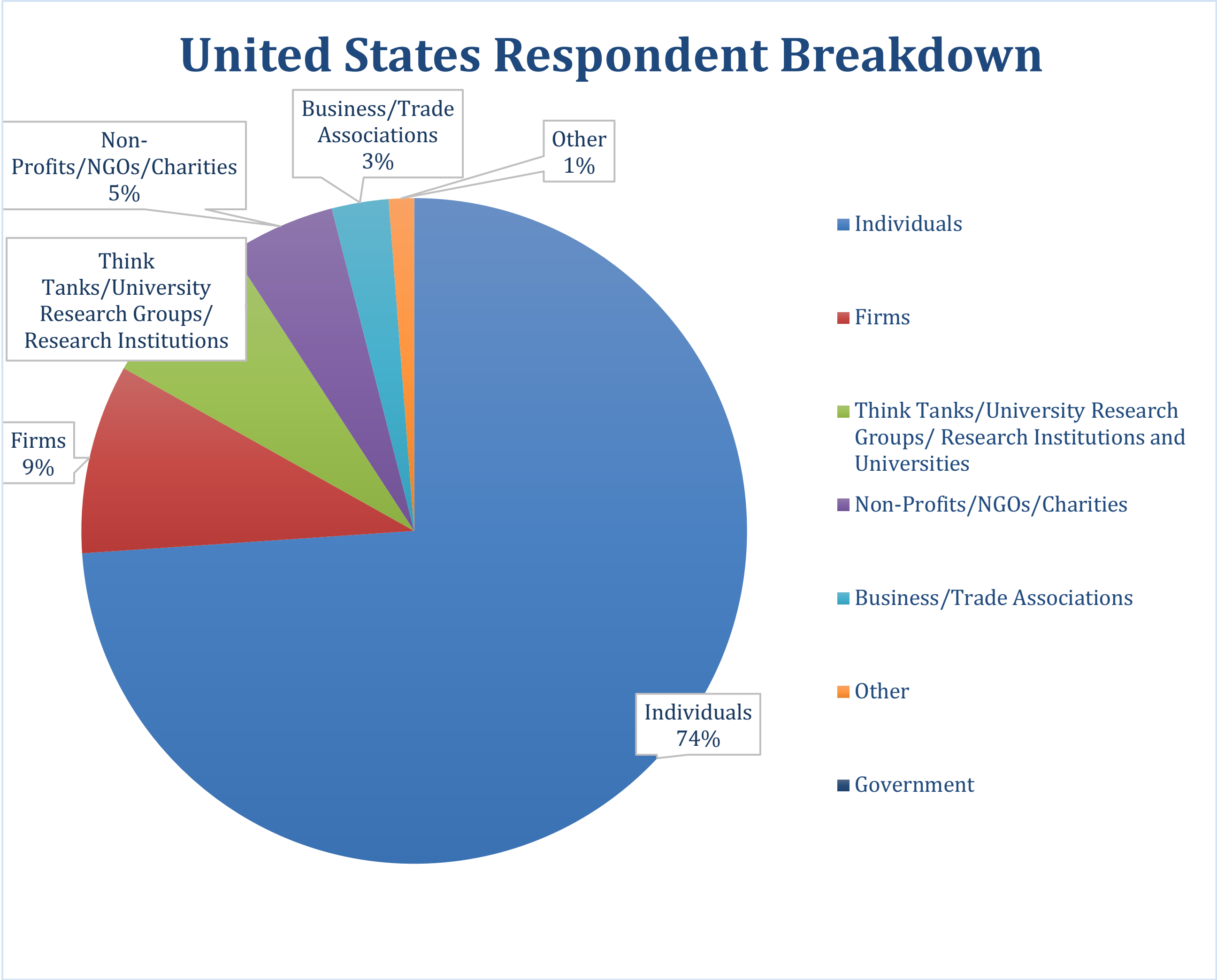

Chart #2 by Michael Moreno

| **NTIA Respondent Breakdown** | | |
|---|---|---|
| **Respondent** | **% of Total** | **Count** |
| Individuals | 73.93% | 241 (81 anonymous) |
| Firms | 9.20% | 30 |
| Think Tanks/University Research Groups/ Research Institutions and Universities | 7.67% | 25 |
| Non-Profits/NGOs/Charities | 5.21% | 17 |
| Business/Trade Associations | 2.76% | 9 |
| Other | 1.22% | 4 |
| Government | 0% | 0 |
| **Total** | | **326** |

Table #3 by Michael Moreno

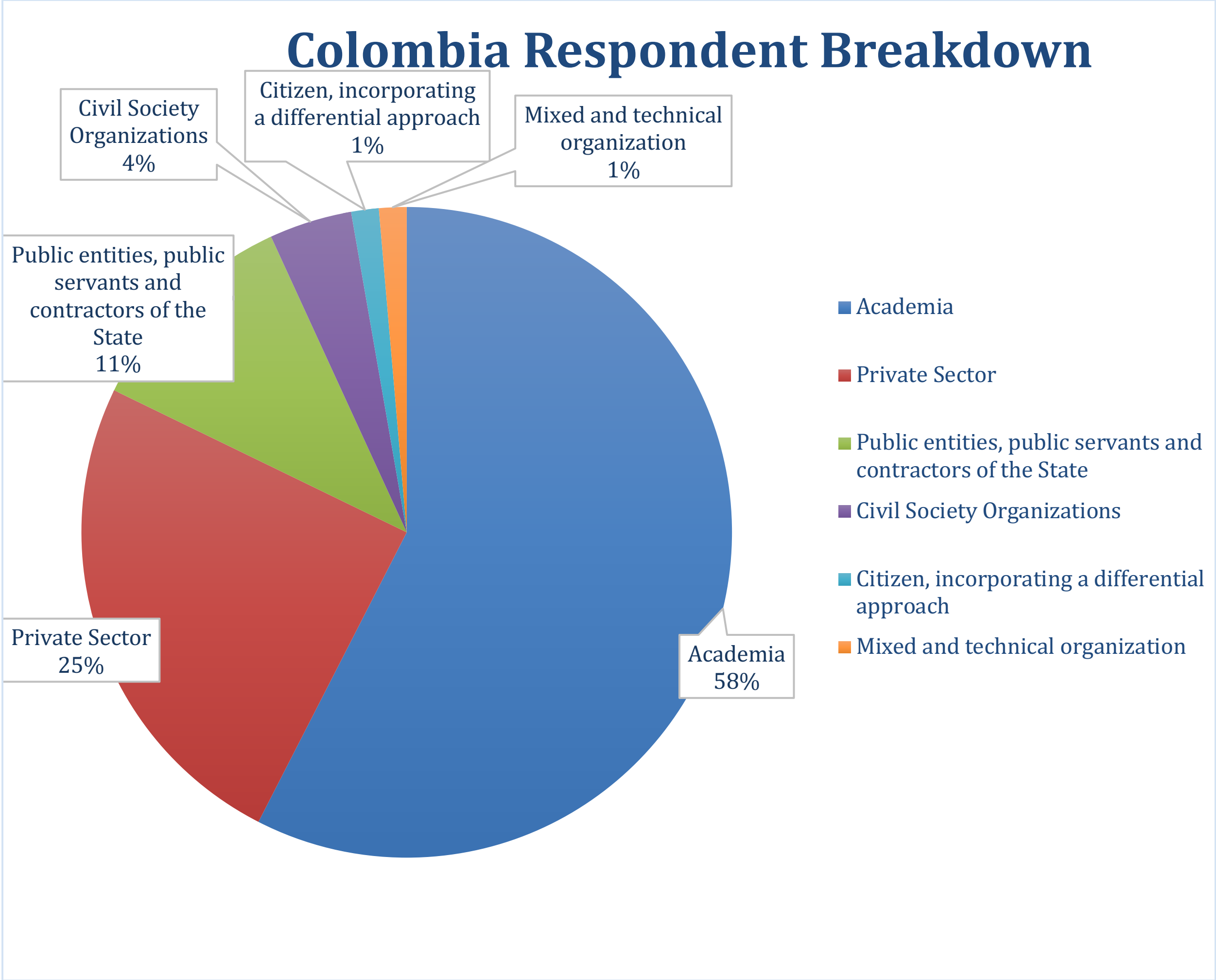

Chart #3 by Michael Moreno

| Colombia Respondent Breakdown | | |
|---|---|---|
| **Respondent** | **% of Total** | **Count** |
| Academia | 57.5% | 42 |
| Private Sector | 24.7% | 18 |
| Public entities, public servants and contractors of the State | 11% | 8 |
| Civil Society Organizations | 4.1% | 3 |
| Individuals | 1.4% | 1 |
| Mixed and technical organization | 1.4% | 1 |
| **Total** | | **73** |

Table #4 by Michael Moreno

The authors found the composition of respondents varied significantly across the three countries,[9] reflecting differences in how the government communicated its request for public input: the outreach strategies, consultation design, and specific public concerns about AI in each country.

**Demographics of Input**

In Colombia, 73 individuals responded to MinCiencias' call for comment. Unlike the other countries, Colombian officials designed a tailored response form for individuals and used it to collect extensive personal demographic information such as ethnicity, disability status, and sex, rather than focusing on organizational affiliation. While some participants indicated they represented firms when describing their experience (notably in question 16), the structure of the data collection made it difficult to reliably match specific comments to either individuals or organizations. Despite these challenges, analysis revealed distinct patterns in sectoral representation and self-identified expertise:

- Sectoral breakdown (Chart #3): Academia dominated responses (57.5%), followed by the private sector/unions (24.7%), public entities (11%), civil society organizations (4.1%), and mixed/technical groups (1.4%).
- Expertise areas (Chart #4): Education, Research, and Innovation (79.5%), Ethics and Governance (39.7%), Innovative and Emerging Industries (34.2%), Data and Organizations (32.9%), and Privacy, Cybersecurity, and Defense (19.2%)

**Colombia Self-Identified Expertise Areas**

[9] Authors' analysis based on respondent composition data from NTIA, Colombia, and Australia public consultations on AI policy

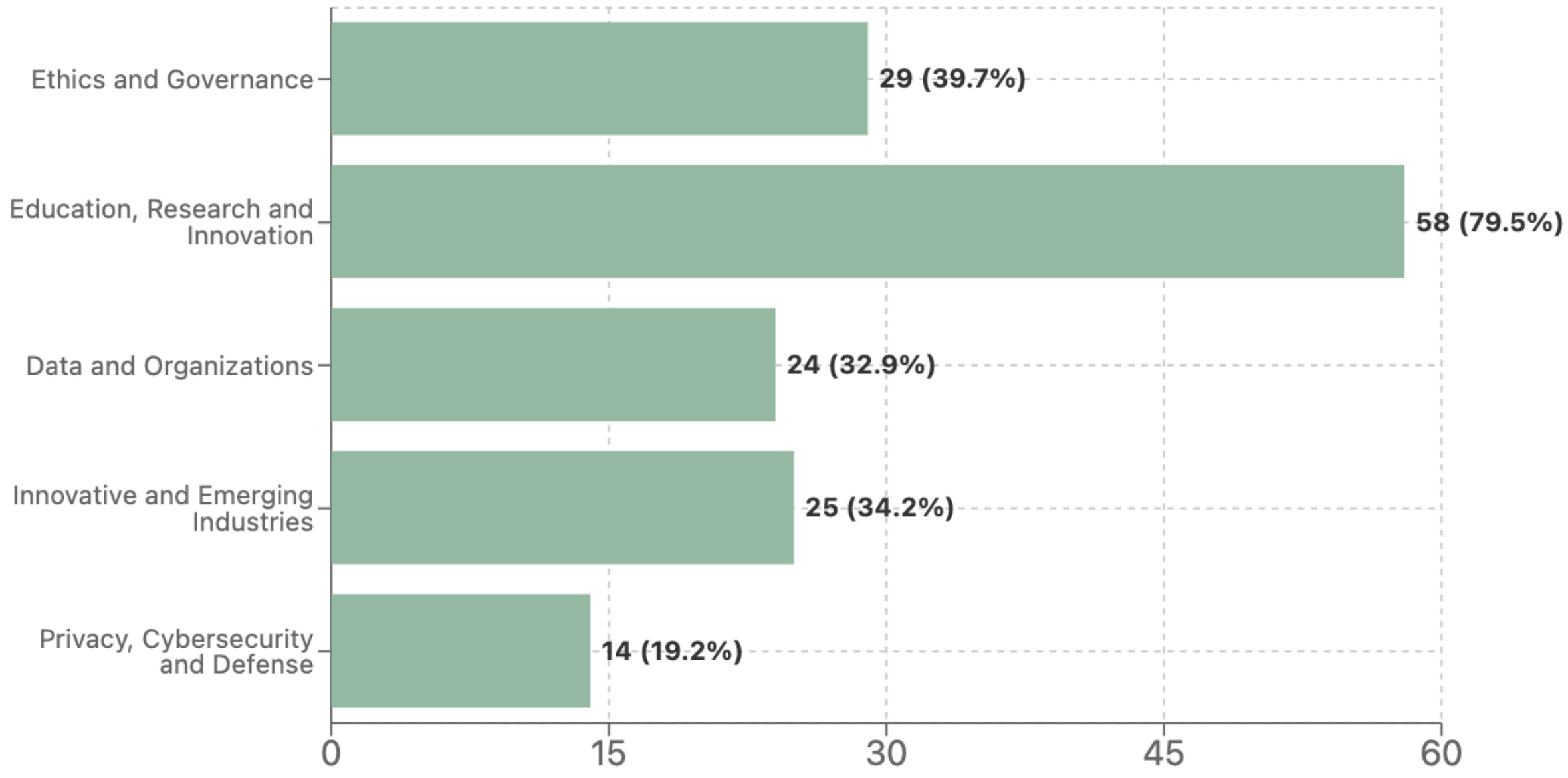

Chart #4 by Michael Moreno (NOTE: Respondents were allowed to click multiple expertise areas.)

This distribution suggests that the consultation in Colombia primarily attracted participants with technical or academic backgrounds, rather than a broad cross-section of society. Policymakers received feedback primarily from experts and institutional stakeholders, rather than from the general public or marginalized groups (MinCiencias 2024c, p.15).

Australia's consultation process attracted the largest response in our sample despite having the smallest population among the three. The Department of Industry, Science and Resources received 510 submissions, with 447 made publicly available as some comments were confidential and submitted privately (Department of Industry, Science and Resources 2023, p.8), reflecting a mix of individual and institutional engagement distinct from both the US and Colombian consultations. While 36% of submissions came from individuals — a stark contrast to the US's 74% — the participation breakdown reveals a more balanced representation of organized stakeholders: Firms (23.70%), nonprofits/charities/NGOs (14.13%), Business/Trade Associations (14.30%), and Think Tanks/University Research Groups/Research Institutions (14.80%). Government entities contributed (2.69%), and 81 of the individual submissions were anonymous.

The consultation process included three ministerial roundtables (64 participants), one virtual town hall (345 participants), four in-person roundtables (59 participants), and four virtual roundtables (81 participants) (Department of Industry, Science and Resources 2023, p.7). However, the authors note they were unable to verify who participated in the live events, as no participant details were released.

In the United States, some 326 individuals or groups submitted comments; Individuals contributed 74% of the commenters', 9% from firms, 5% from nonprofits/NGOs, 8% from think tanks or research institutions, and 1% from other entities. Notably, 34% of individual respondents (81 individuals) chose anonymity, though some partially disclosed professions (e.g., "artist," "systems engineer") to contextualize their expertise (Aaronson: 2025 p. 11). Interestingly, anonymous contributors strongly advocated for open-source AI systems. In the case of Colombia, the government's consultation form[10] specifically prompted them to describe their work sector, areas of expertise, and relevant experience (notably in question 16).

In sum, none of the three governments were able to engage a broad section of their populace.

**2. What materials did the government provide to prepare the public to give informed advice?**

The authors found the three governments provided some material to ensure that their constituents had baseline knowledge to comment on AI policies. The NTIA included a "Supplemental Information" section with its call for comment, providing background, the legal authority for the consultation, definitions of key terms, and a comprehensive list of 9 major questions with 52 sub-questions. Even the most engaged would struggle to answer all of these questions in an in-depth manner, and we found most only answered a few questions. However, many of the participants did not answer the bulk of the questions, and some seemed confused by the specifics and sheer number of the questions (Aaronson: 2025).

In contrast, Colombia's Ministry of Science, Technology, and Innovation released a preliminary version of its AI Roadmap just two days before the public consultation opened (MinCiencias: 2024a). The brief consultation window—from January 29 to February 6, 2024—combined with limited public outreach, constrained both awareness and participation, resulting in only 73 individuals taking part. Citizens were notified by the government via a post on their official website and social media that the paper would be released in January 2024 (MinCiencias: 2023a; MinCiencias: 2023b). The roadmap identified their five main focus areas: Ethics and Governance; Education, Research, and Innovation; Data and Organizations; Innovative and Emerging Industries; Privacy, Cybersecurity, and Defense (MinCiencias: 2024a, p. 42-43, 45-46, 48). While the roadmap provided a comprehensive outline and sustainable and ethical considerations. MinCiencias did not make it easy to find the draft, as they did not include the consultation document in its call for comment on the website, or social media. Therefore, access to the draft may have been limited, one respondent noted difficulty even finding the document.[11]

[10] "Consulta Ciudadana de Hoja de ruta para garantizar la adopción ética y sostenible de la Inteligencia Artificial en Colombia," Google Forms, https://docs.google.com/forms/d/e/1754886223657623598.
[11] Refer to commentor #47; https://tinyurl.com/4j5erjay

The government administered the consultation in a Google Form that included both technical and personal questions. However, because the consultation window was short and the officials did not attach the consultation document to the call for comment on their website and social media, the broad public was unlikely to hear about it, let alone participate. Moreover, many participants did not answer all the questions.[12] Hence, Colombian policymakers did not allow enough time and sufficiently prepare the public for meaningful engagement.

Australia offered a range of materials to help citizens better understand the issues and provide informed feedback. The government published a discussion paper at the start of the consultation on June 1, 2023 (Department of Industry, Science and Resources 2023), along with an overview of the consultation as well as supplemental documents labeled "read more" on the official website, which offered related information.[13] It also included a section "related" section linking to previous consultations enabling respondents to review the consultation process (Department of Industry, Science and Resources, 2022).  With this approach, the government was able to have citizens comment at different times and in different ways. However, the materials assumed that many respondents would have a baseline familiarity with AI governance, safety, and risk concepts. Despite these resources, participation data showed that not all respondents answered every question, suggesting persistent accessibility gaps for non-expert stakeholders.[14]

To summarize, while all three governments provided some materials on the background, and objectives or the consultation, these documents alone did not equip the public to give well-informed comments on AI safety and risk.

**3. Did policymakers make efforts to ensure a broad cross-section of people knew about and could comment on the proposed policy?**

---

[12] The author's spreadsheet delineating this analysis will be placed on the research section of the Digital Trade and Data Governance Hub website so that individuals can review the data set. See https://datagovhub.elliott.gwu.edu/research-overview/.

[13] The Department of Industry, Science and Resources offered three documents labeled "read more":https://www.industry.gov.au/science-technology-and-innovation/technology/artificial-intelligence;https://www.chiefscientist.gov.au/GenerativeAI;https://www.industry.gov.au/science-technology-and-innovation/technology

[14] The author's spreadsheet delineating this analysis will be placed on the research section of the Digital Trade and Data Governance Hub website so that individuals can review the data set. See https://datagovhub.elliott.gwu.edu/research-overview/.

| Overview of Public Consultation Timelines | | | | |
|---|---|---|---|---|
| Country | Date | Phase | Key Event/Action | Documents/Output |
| United States (NTIA) | October 30, 2023 | 1 | President Joe Biden issues Executive Order mandating NTIA to gather public comments on AI-related matters | President Biden Executive Order #14110 |
| | December 13, 2023 | 2 | Alan Davidson announces the call for comments during speech at Center for Democracy & Technology (CDT) | CDT Speech |
| | February 20, 2024 | 3 | Call for comments formally posted on Federal Register, opening public comment period | Federal Register Notice, including background materials and detailed question list |
| | March 27, 2024 | 4 | NTIA closes the public comment period | - |
| | July 2024 | 5 | NTIA releases report to the White House | Final Report to White House |
| Colombia (MinCiencias) | August 11, 2023 | 1 | Annouced the creation of a Roadmap for the development and application of AI in Colombia. | During the forum "The Transformative Potential of AI" and in a post on their website MinCiencias annouced a roadmap to be created. |
| | December 24, 2023 | 2 | On their website MinCiencias the roadmap will be released in January 2024. | - |
| | January 29,2024 | 3 | Call for comments formally posted on Google Forms, marketed on their social medias and websites. | The draft roadmap was released Janaury 27, 2024 however their was no offical posting their website of this draft. |
| | February 6,2024 | 4 | Call for comment closed | Public responses available |
| | February 12, 2024 | 5 | Final versiona of the roadmap was released | Final roadmap released |
| Australia (DISR) | June 1, 2023 | 1 | Call for comment opened | Consultation document was released, along with other background and supporting documents |
| | August 4, 2023 | 2 | Call for comment closed | - |
| | January 17, 2024 | 3 | Response to the public publsihed | Interim Response published by DISR |

Table #5 by Michael Moreno

Democracies seek public comment on a wide range of issues from taxes to environmental government. It is not easy to get people to comment, and it is even harder to get a representative sample of citizens to comment (OECD: 2025a). Hence policymakers must find ways to inform their citizens that they seek their input on such specific issues by marketing the calls to various key constituencies. Yet, none of the 3 governments' cases really tried to “market” the consultations to their citizens. In the United States, the Biden Administration announced a call for public comment on open source in President Biden's executive order on AI. Then Assistant Secretary Alan Davidson announced the call at several Washington area think tanks. The NTIA also issued a Federal Register notice, allowing responses online, by mail, and by telephone, and provided contact information for staff to answer questions. However, there is little evidence of targeted outreach beyond these speeches and formal channels (Aaronson: 2025). The NTIA neither publicized roundtables nor partnered with civil-society organizations, and it provided only a brief comment period. Consequently, while its process remained transparent and accessible, the respondents were insiders and not the broad public.

All 3 governments relied heavily on government websites to request and collect public comments on AI safety and risk. Hence in order to comment, one needs internet access, which is not available or affordable for all people. For example, MinCiencias announced the consultation via a post on its official website and social medias, with a very brief window—just over a week—for public comment (MinCiencias: 2024e; MinCiencias: 2024c). The post included a link to the Google Form to collect comments, the closing date and noted the importance of citizen involvement (MinCiencias: 2024c).[15] This short timeline (the call for comment opened on 1/29/24 and closed on 2/6/24 (MinCiencias: 2024c) and minimal publicity likely restricted awareness and participation, with them receiving only 73 individuals participating in the consultation, suggesting little effort to engage a wide cross-section of society. Additionally, the MinCiencias announced the call on their social media, however, this had been posted February 1st, just five days before the call closed (MinCiencias: 2024e). The authors could find no evidence that the Colombian government undertook additional outreach measures to receive public comment, such as partnerships with civil society organizations, media coverage, or community events, to broaden awareness beyond its immediate digital audience.[16] To confirm, the authors had reached out to MinCiencias requesting any information they had regarding how they used the responses and if any media campaigns were conducted. The ministry responded by stating:

"The comments, observations, and citizen proposals received during the Citizen Consultation were carefully analyzed by the technical team of the Directorate of Technological Development and Innovation. This analysis identified relevant contributions from various sectors of society, which were incorporated into the final document as part of an exercise to refine the initial proposals. The consultation sought to validate and enrich aspects such as 1. The proposed innovation paths, 2. The identification of strategic problems, 3. The stated objectives, 4. And the priority innovation focuses. The call for proposals was conducted through the Ministry of Science's official communication channels, including our institutional accounts on Facebook, Instagram, Twitter, and the official website, where a public link was provided to participate. The invitation was also extended to various state entities and public actors, seeking their perspectives from their institutional roles and responsibilities."[17] However, the Colombian government had already consulted academics, industry leaders, and other key stakeholders (MinCiencias: 2024d, p16).

---

[15] Authors note that the Colombia government made a post via their website and announcement at the forum 'The transformative potential of Artificial Intelligence: challenges in the field of ethics and biodiversity conservation' , the Ministry of Science, Technology and Innovation announced the creation of a roadmap in August 2023. As well, in December 2023, the Ministry of Science stated they will launch the 'Roadmap to ensure the ethical and sustainable adoption of Artificial Intelligence in Colombia' in January 2024.

[16] Based on author reaching out to MinCiencias for comments, and reviewing the consultation websites, government social media posts, and major news outlets between April and June 2025. No evidence of broader outreach was identified.

Australia similarly relied on government websites and forms in their consultation process. However, policymakers did more than the US to inform its citizens about the consultation. After issuing a discussion paper, the government conducted a two-month consultation period, generating 510 submissions online and conducted a combination of 3 ministerial roundtables, 1 virtual town hall, and 8 additional in-person and virtual roundtables. The authors note that this expanded engagement by the government—combining ministerial roundtables, a virtual town hall, and multiple in-person and online sessions—was designed to get a broad sample of Australians. Additionally, the authors note that this public consultation directly overlapped with the release of the Royal Commission's report into the *RoboDebt Scheme*—a controversial automated debt recovery program that unlawfully used algorithmic tools to identify and reclaim welfare overpayments (Royal Commission into the Robodebt Scheme: 2023). The scandal, which disproportionately affected vulnerable populations, may have increased the public's willingness to participate in AI-related consultations.

To summarize, the U.S. NTIA launched its AI consultation through speeches and a Federal Register notice but did not put much effort into encouraging discussion beyond the usual respondents—people who had a stake in open vs. closed source issues. Colombia's one-week, single-post consultation on MinCiencias' website drew only a handful of responses, indicating limited public awareness and minimal effort to broaden participation. By contrast, Australia's eight-week "Safe & Responsible AI" process paired an online discussion paper with ministerial roundtables and a virtual town hall, ensuring a more diverse cross-section of citizens could comment. So, each nation tried to get public engagement through varied mechanisms and with varied degrees of effort. But none of the three governments really made a sustained effort to encourage broad participation.

**4. Did the governments provide evidence that it used the feedback it received?**

[17] Email to Michael Moreno on May 19, 2025, from MinCiencias responding on inquiries on the consultation. In Moreno's files.

If officials want to sustain public legitimacy and trust, they must not only acknowledge but truly listen to what their constituents have say. Such a government is more likely to create an effective feedback loop between the government and those it governs (OECD: 2011). The authors therefore evaluated how each of the 3 case studies reported on and utilized the comments it received.

In contract with other calls for public comment, the US did not provide many specifics regarding who commented and what they said. In a March 21, 2024, speech to the Center for Strategic and International Studies (CSIS), NTIA Assistant Secretary Alan Davidson noted that the agency had "listened to concerns" raised during its public consultation on open-source AI systems. But he did not discuss what they heard. Moreover, in the NTIA report on open-source AI risks, the government noted it had sought comment but did not share it or show how these comments influenced the report (NTIA: 2024, p.3 as cited by Aaronson 2025).

Colombia's government asserted, in their final roadmap, that citizen feedback "improved" its AI Roadmap by incorporating "opinions, comments, technical suggestions, and proposals for new innovation paths." However, the final document, released on February 12, 2024, only 6 days after the consultation closed, retained the structure and content of the original draft without highlighting modifications derived from the consultation (MinCiencias: 2024d, p. 16).

After the June–August 2023 consultation, the Australian government released an interim report on January 17, 2024, detailing how the government engaged with citizens, what they heard, and the government's interim response (Department of Industry, Science and Resources: 2024a). In response to explicit public comment, the document mapped public concerns to actionable measures, including establishing a "Temporary Expert Advisory Group" to design AI guardrails, responding to submissions urging specialized oversight and committing to explore a permanent advisory body, reflecting widespread stakeholder demand for sustained expert-public collaboration (Department of Industry, Science and Resources: 2024a, p. 21). The government s responded to suggestions from the US-based company Salesforce (submission #147); an anonymous individual (#398), and the Insurance Council of Australia (submission #103), among others.

In summation, we were most struck by the failure of all 3 governments to be truly responsive to citizen concerns.

## 5. Applying the IAP2 Spectrum of Political Participation

Lastly, the authors wanted to gauge how much effort policy makers put into meaningful public involvement. To do that, we turned to IAP2's Spectrum of Public Participation— which illustrates different modes of citizen participation—and supplemented it with findings from our core research questions. This allowed us to characterize if a nation's engagement strategy for the development of AI governance went beyond consulting the public, to more actively involving them in the process.

| Draft IAP2 Mapping | | | | | | | |
|---|---|---|---|---|---|---|---|
| **Countries** | **Inform** | **Consult** | **Acknowledge** | **Responsive** | **Involve** | **Collaborate** | **Empower** |
| **United States** | ✓ | ✓ | ✓ | | | | |
| **Colombia** | ✓ | ✓ | ✓ | | | | |
| **Australia** | ✓ | ✓ | ✓ | | | | |

Table #6 by Michael Moreno

As the table shows, the United States, Colombia, and Australia all took steps to inform, consult, and acknowledge the public, but did little to involve, collaborate with, or empower stakeholders throughout the process. Each government provided background information and explanatory materials to help the public understand the issues prior to publishing their strategies. They also solicited comments from both organized and unorganized publics—through formal notices, online forms, public forums, and, in Australia's case, a series of roundtables and town halls. All three governments acknowledged receipt of public input: the US referenced comments in official speeches and documents (NTIA: 2024, p. 3), Colombia described public feedback as "improving" its final draft (MinCiencias: 2024d, p. 17), and Australia published an "Interim Response" report summarizing stakeholder perspectives and how the government will ensure AI is designed, developed and deployed safely and responsibly (Department of Industry, Science and Resources: 2023, p. 5-6).

However, none of the 3 governments really took the comments they received and changed policies. Hence, we cannot describe any of these governments as "responsive." Australian officials explained how the feedback influenced the creation of a temporary expert advisory group, which operated till September 30, 2024 (Department of Industry, Science and Resources: 2024b)). In contrast, neither the US nor Colombia provided documentation showing how officials heard and responded to public feedback.

These three cases led us to believe that the current consultative process is not working.
It is not effectively marketed, officials don't give their citizens the materials they need to fully participate, and often policymakers ignore, downplay or fail to utilize what citizens told them.

**Final Thoughts / Recommendations**

Democracy must be a two-way street. Policymakers should seek, listen to, and utilize public comment (Sheldon: 2023) If public concerns are lost in translation, policymakers may find that the public views public policies as unworthy of their trust. Unfortunately, in a recent study, the OECD found that on average across OECD countries, only 32% of citizens find it likely that the government would adopt the opinions expressed in a public consultation (OECD: 2025, p. 17). The 2024 OECD Trust Survey found on average, across OECD countries, only 39% of people have high or moderately high trust in their national governments.

Policymakers need to reassure the public that their views matter (OECD: 2024b). In general, policymakers have relied on in-person meetings or online public comment portals to seek such comment. But those strategies are not encouraging the public to comment.[18] Instead policymakers could:

1. **Support civic and digital literacy** so that individuals can better understand AI technologies, how they relate to their lives, and how to contribute effectively to public consultations. Moreover, policymakers should provide tailored and easy to understand background materials on AI. Such materials could include video explainers, primers, and white papers. In addition, policymakers could create a centralized clearly accessible repository of relevant background materials.[19]

2. **Establish, maintain, and consistently monitor public comment portals** to demonstrate that public opinion is regularly sought and valued. Policymakers should be specific and responsive to their constituents if they want to sustain trust in their governance of AI. Officials should establish a clear point of contact who is always available to hear or receive feedback and policymakers should show how they responded.

3. **Widely market calls for comment** through both social and traditional media, using trusted voices—such as celebrities, academics, business leaders, and policymakers—to encourage broad participation.

4. **Regularly hold online town halls** on AI across the country and ensure that public concerns are genuinely heard and addressed.

5**. Utilize innovative engagement strategies** like policy hackathons, public challenges, and crowdsourcing initiatives to bring in diverse perspectives.

---

[18] See as example, General Services Administration, how members of the public can contribute to the regulatory process, https://www.gsa.gov/policy-regulations/regulations/managing-the-federal-rulemaking-process-erulemaking/how-members-of-the-public-can-contribute-to-the-re#:~:text=Why%20are%20public%20comments%20important,the%20outcomes%20of%20the%20regulations

[19] We are grateful to one of our anonymous reviewers for this suggestion.

6. **Ensure participation from underrepresented groups**, including those without reliable internet access or from marginalized socioeconomic or ethnic communities, by designing inclusive consultation processes.

**7. Be responsive to what people say**. They can help policymakers be agile and anticipatory because they may perceive things differently from typical advisors from business or academia.

Ironically, various AI systems can help policymakers update these processes by involving more people in a short period of time and helping them to feel heard.

For example, variants of swarm AI can enable large numbers of people to deliberate and find consensus. A recent study by Carnegie Mellon researchers used Artificial Swarm Intelligence (ASI), which allows networked groups of people to make collaborative decisions by deliberating in systems modelled on biological swarms such as flocks of birds. In 2023, researchers combined ASI with large language models. The researchers worked with two groups of 75 participants who took part in a 30-minute session in which they brainstormed using large language models and in traditional chat room). Then, each individual completed a subjective feedback survey to compare the two experiences. The authors found that:

- Participants significantly preferred using ASI reporting that it felt more collaborative and more productive and was better at producing high-quality answers.
- More than 80% of the participants reported feeling "more heard" during each CSI deliberation and came away feeling "more ownership" and "more buy-in" for the resulting answers than they did in a traditional real-time chat environment. The process can work with groups larger than 250 people[20]

Moreover, AI-powered technologies can help increase the impact of participatory processes by helping governments analyze and make sense of large amounts of data from their constituents. Large language models can analyze and summarize large pools of data in a short period of time (OECD: 2025b). Finally, generative AI can lower the barriers to participate by helping the public navigate complex or technical language and aid participants using free chat-bots such as Chat-GPT or Gemini.

In sum, policymakers need to rethink when and how they will engage their citizens in governing AI. Policymakers need citizens to comment and citizens as well as policymakers don't want their comments on AI to be lost in translation. With guardrails, AI could be a tool to help facilitate that engagement, and in so doing sustain democracy.[21]

[20] https://www.cmu.edu/tepper/news/stories/2025/april/using-principles-of-swarm-intelligence-study-compared-platforms-that-allow-brainstorming-among-large-groups.html?session_id=session-c0wfjshx4

[21] https://www.theguardian.com/technology/2025/jun/15/government-roll-out-humphrey-ai-tool-reliance-big-tech